\documentclass[10pt,twocolumn,letterpaper]{article}

\usepackage[pagenumbers]{wacv} 

\usepackage{graphicx}
\usepackage{amsmath}
\usepackage{amssymb}
\usepackage{booktabs}

\usepackage{multirow}
\usepackage{bbding}
\usepackage{pifont}
\usepackage{xcolor}
\usepackage{colortbl}
\usepackage{graphicx}
\usepackage{subcaption}
\usepackage[accsupp]{axessibility}  

\usepackage[pagebackref,breaklinks,colorlinks]{hyperref}

\usepackage[capitalize]{cleveref}
\crefname{section}{Sec.}{Secs.}
\Crefname{section}{Section}{Sections}
\Crefname{table}{Table}{Tables}
\crefname{table}{Tab.}{Tabs.}

\def\wacvPaperID{427} 
\def\confName{WACV}
\def\confYear{2024}

\begin{document}

\title{From Denoising Training to Test-Time Adaptation: \\ Enhancing Domain Generalization for Medical Image Segmentation}


\author{Ruxue Wen\hspace{.1in} 
        Hangjie Yuan\hspace{.1in} 
        Dong Ni\thanks{Corresponding author.}\hspace{.1in} 
        Wenbo Xiao\hspace{.1in} 
        Yaoyao Wu\\
Zhejiang University\\
Hangzhou, Zhejiang, China\\
{\tt\small \{ruxue.wen, hj.yuan, dni, xiaowenbo, yaoyaowu\}@zju.edu.cn}
}

\maketitle

\begin{abstract}
    In medical image segmentation, domain generalization poses a significant challenge due to domain shifts caused by variations in data acquisition devices and other factors. These shifts are particularly pronounced in the most common scenario, which involves only single-source domain data due to privacy concerns.
    To address this, we draw inspiration from the self-supervised learning paradigm that effectively discourages overfitting to the source domain.
    We propose the Denoising Y-Net (DeY-Net), a novel approach incorporating an auxiliary denoising decoder into the basic U-Net architecture.
    The auxiliary decoder aims to perform denoising training, augmenting the domain-invariant representation that facilitates domain generalization.
    Furthermore, this paradigm provides the potential to utilize unlabeled data.
    Building upon denoising training, we propose Denoising Test Time Adaptation (DeTTA) that further: \textbf{(i)} adapts the model to the target domain in a sample-wise manner, and \textbf{(ii)} adapts to the noise-corrupted input.
    Extensive experiments conducted on widely-adopted liver segmentation benchmarks demonstrate significant domain generalization improvements over our baseline and state-of-the-art results compared to other methods.
    Code is available at \url{https://github.com/WenRuxue/DeTTA}.
\end{abstract}

\section{Introduction}
\label{sec:intro}
In the last decade, deep learning has been extensively studied for assisting medical image analysis, aiming to reduce doctors' workload.
Medical image segmentation, a critical prerequisite for various clinical analyses, has received significant attention. 
Many deep neural networks~\cite{litjens2017survey}, represented by U-Net~\cite{ronneberger2015u}, demonstrate remarkable performance in various medical image segmentation tasks. 
However, in clinical practice, medical images often display distributional discrepancies due to factors such as different equipment, diverse imaging parameters, and fluctuations in signal-to-noise ratio over time. 
While cross-modality datasets exhibit greater domain shift (e.g., from MRI to CT), it is important to note that scenarios involving the same modality are more prevalent in clinical practice. Therefore, we restrict our focus solely to discrepancies in cases involving the same modality.
Such discrepancies challenge deep neural networks to generalize to unseen domains, leading to performance degradation. 

To address this problem, domain generalization (DG) has emerged to enhance the generalization ability of deep neural networks to unseen domains.
Most existing domain generalization methods~\cite{li2018domain, zhou2020domain, zhu2021self, peng2019moment} aim to achieve generalization performance in unseen domains by extracting domain-invariant features from multiple domains~\cite{liu2022single}. 
These methods prove ineffective when dealing with the problem of medical image segmentation due to the scarcity of available data and available data annotations~\cite{liu2022single}.

In medical image segmentation, a more realistic yet challenging setting is single domain generalization (SDG), where only one single domain is available for training.
For the challenging SDG problem, an intuitive solution is to increase the diversity of training data through adversarial data augmentation~\cite{volpi2018generalizing,zhou2020deep,qiao2020learning} or data generation~\cite{rahman2019multi,somavarapu2020frustratingly,zhou2020learning,li2021progressive}.
However, synthesizing high-quality medical images with intricate details is challenging, and these methods often struggle to perform well in domains that differ significantly from the source domain due to the challenge in anticipating the distribution of test data~\cite{liu2022single}.
In addition to data manipulation, SDG is also studied in general machine learning paradigms~\cite{wang2022generalizing}, such as dictionary learning~\cite{liu2022single}, contrastive learning~\cite{hu2023devil,kim2021selfreg,duboudin2021encouraging}. 
Furthermore, there are also several studies embarking on the exploration of leveraging self-supervised learning, such as predicting the shuffling order of patch-shuffled images \cite{noroozi2016unsupervised} or rotation degrees \cite{gidaris2018unsupervised}, to enhance domain generalization performance. 
An intuitive explanation is that the self-supervised learning paradigm allows a model to learn generic features and reduces the likelihood of overfitting to the source domain~\cite{Carlucci_2019_CVPR}. 

Inspired by the success of self-supervised learning for generalization, we aim to address the SDG problem in a novel way for medical image segmentation. We observe that medical images often suffer from various types of noise due to limitations in imaging technology or variations in imaging protocols, which is one of the key factors contributing to domain shift~\cite{karani2021test}.
Hence, properly leveraging self-supervised denoising can disregard the noise in medical images from different domains, allowing the network to focus more on clean images. 
Secondly, self-supervised denoising benefits from all available raw images, thereby enhancing the feature extraction capabilities of the encoder\cite{buchholz2021denoiseg}. 
Thirdly, the single given test data hints at its distribution, enabling us to adapt the model to each unlabeled test data at test time only. This test-time adaptation approach is compatible with solving the SDG problem~\cite{liang2023comprehensive}.

With these insights, we present Denoising Y-Net (DeY-Net), a novel approach with a Y-shaped architecture to enhance domain generalization for medical image segmentation. 
DeY-Net consists of an encoder followed by two decoders: a decoder for pixel-wise segmentation and an auxiliary decoder for self-supervised denoising, utilizing the Noise2Void training scheme~\cite{krull2019noise2void}. 
Furthermore, the self-supervised denoising branch provides the potential to utilize unlabeled data.
Building upon denoising training, we propose Denoising Test Time Adaptation (DeTTA) that further: \textbf{(i)} adapts the model to the target domain in a sample-wise manner, and \textbf{(ii)} adapts to the noise-corrupted input, achieving more stable performance improvements.

Our main contributions are highlighted as follows:
\begin{itemize}
    \item 
    We present a novel architecture named DeY-Net, to address the SDG problem by incorporating a self-supervised denoising decoder into a basic U-Net. 
    \item 
    We propose Denoising Test-Time Adaptation (DeTTA), which adapts the model to the target domain and adapts to the noise-corrupted input in order to preserve more information.
    \item 
    We conduct extensive experiments on a widely-adopted liver segmentation task. By training only on a single domain, our method significantly improves generalization performance over our baseline and state-of-the-art results compared to other methods.
\end{itemize}

\section{Related work}

\subsection{Denoising for segmentation} \label{denoise}

It is well-known that there is an overlap between denoising and segmentation tasks~\cite{zamir2018taskonomy}. 
Various works have proved that the self-supervised denoising task can enable the segmentation task, especially in the presence of extreme levels of noise and limited training data~\cite{prakash2020leveraging}. Mangal Prakash \etal~\cite{prakash2020leveraging} demonstrate that the self-supervised denoising  prior~\cite{krull2019noise2void} can significantly improve segmentation results, where denoising and segmentation are realized in two sequential steps. Similarly, Sicheng Wang \etal~\cite{wang2019segmentation} use a network that incorporates tandem segmentation and denoising tasks. Tim-Oliver Buchholz \etal~\cite{buchholz2021denoiseg} propose using a single network to jointly predict the denoised image and the desired object segmentation. Emmanuel Asiedu Brempong \etal~\cite{brempong2022denoising} propose Decoder Denoising Pretraining (DDeP) to pretrain the segmentation decoder with a well-trained denoising network. These approaches utilize self-supervised denoising to improve segmentation performance via network architecture or training schemes.
Building on this inspiration, we present a novel Y-shaped architecture integrating self-supervised denoising as a secondary decoder in the network.

\subsection{Self-supervised learning} \label{self}

Self-supervised learning (SSL) is a learning paradigm that enables learning semantic features by generating supervisory signals from a pool of unlabeled data~\cite{shurrab2022self}. In the medical field, several works have demonstrated that SSL can produce a pretrained model to advance supervised tasks~\cite{huang2023self}, such as image classification~\cite{google-research-blog,huang2023self} and image segmentation~\cite{kalapos2022self,ouyang2022self}.
These approaches learn image representations through handcrafted pretext tasks~\cite{shurrab2022self} such as image rotation prediction~\cite{gidaris2018unsupervised}, in-painting\cite{pathak2016context}, Jigsaw puzzle~\cite{noroozi2016unsupervised}, denoising auto-encoder~\cite{vincent2008extracting} and so on.
Besides, for generalization purposes, some methods utilize self-supervised learning as an auxiliary task for the main task~\cite{Carlucci_2019_CVPR}.
As early as 2016, Muhammad Ghifary \etal~\cite{ghifary2016deep} propose to add a reconstruction decoder that shares the encoding representation with the classification head, which can be trained with unlabeled target domain data. Inspired by this, Joris Roels \etal~\cite{roels2019domain} propose a domain adaptation (DA) method, named Y-Net, by integrating a reconstruction decoder for medical image segmentation within the Y-shaped architecture. Y-Net is also proposed in ~\cite{wang2019net}. The distinction between DA and DG lies in their utilization of target domain data during the training phase, which is exclusive to DA and not employed in DG. Kai Zhu \etal~\cite{zhu2020self} devise a Self-Supervised Module (SSM) to improve the segmentation performance. Yu Sun \etal~\cite{sun2020test} integrate a self-supervised image rotation classifier head, allowing for the utilization of the unlabeled test data at test time.
Inspired by these works, we further explored the effectiveness of incorporating a self-supervised branch to enhance the model's generalization performance.

\begin{figure*}[th]
\vspace{-1mm}
\includegraphics[width=1\linewidth, trim=0cm 2cm 0cm 2cm, clip]{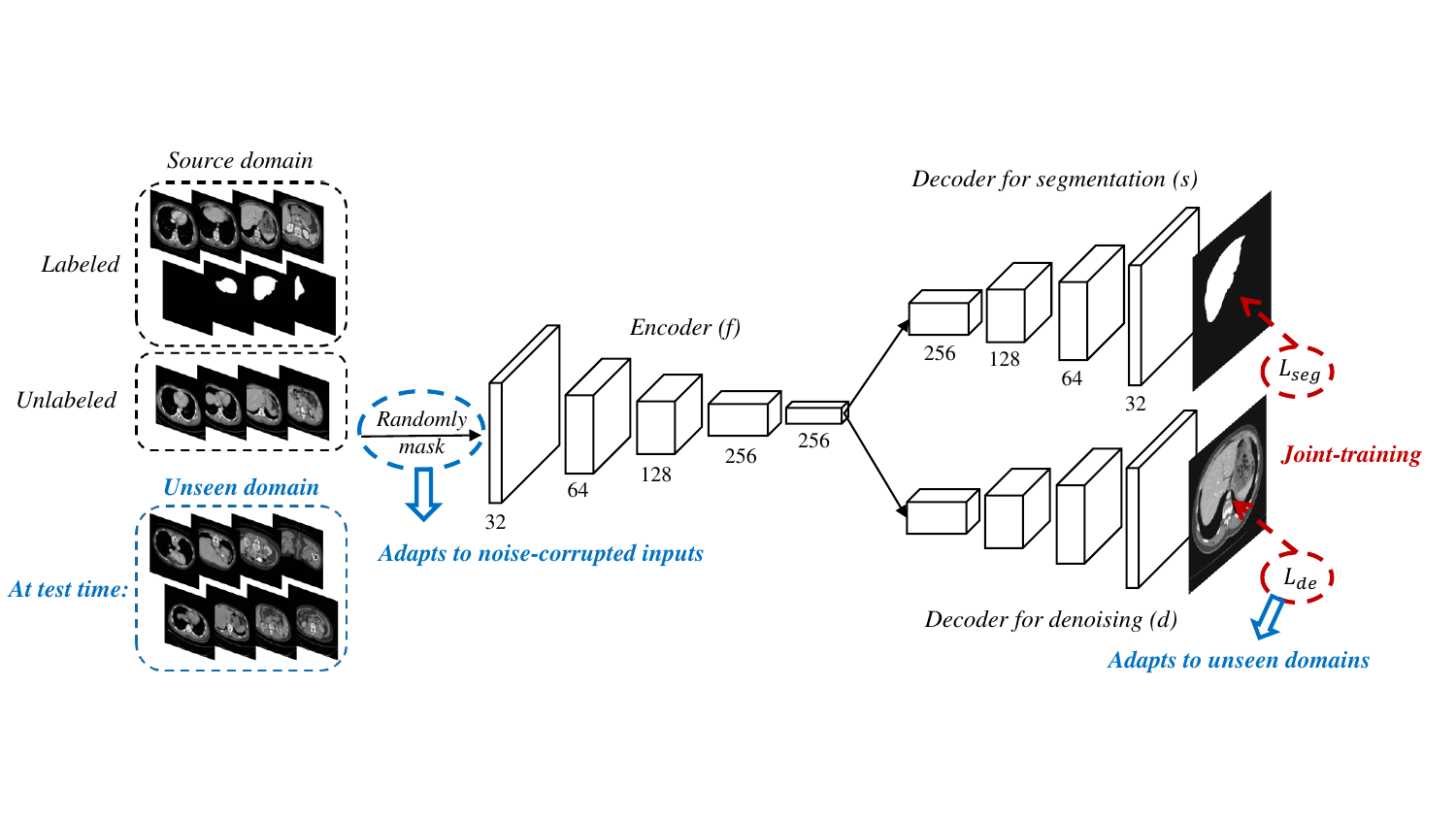}
\vspace{-8mm}
   \caption{
   Overview of the proposed DeY-Net.
   The fundamental backbone is U-Net, and the structures of the two decoders are identical and retain the skip connections (omitted in the figure).
   We joint-train the segmentation and the self-supervised denoising tasks by combining their losses $L_{seg}$ and $L_{de}$. 
   When testing data on a new domain, we only adapt the encoder parameters through the self-supervised denoising branch with $L_{de}$ to improve generalization. 
   Besides, each test image is randomly masked with neighbor pixels several times to get a more stable average prediction.}
\label{fig:Overview}
\vspace{-4mm}
\end{figure*}

\subsection{Test-time adaptation}
In recent years, there has been significant development in Test-Time Adaptation (TTA). 
TTA aims to utilize the distribution information from the test data to quickly adapt models with a few gradient steps~\cite{pandey2021generalization,chen2021source,liu2022single}.
The main differences of prior work lie in how to devise the objective that can be optimized with unlabeled test data and which part of network parameters to be updated at test time~\cite{liu2022single}. 
For instance, TTT~\cite{sun2020test} adapts the encoder of the classification model at test time via an auxiliary branch with rotation prediction self-supervision. Later on, Tent~\cite{wang2020tent} optimizes entropy minimization loss of predictions on test data to adapt the batch normalization layer. 
For test-time adaptation on medical image segmentation, Hu \etal \cite{hu2021fully} propose using new losses like Regional Nuclear-norm (RN) and Contour Regularization (CR) losses to improve generalization performance.
Neerav Karani \etal ~\cite{karani2021test} propose to generate pseudo-labels through a denoising autoencoder at test time to adapt an image normalization module. The denoising autoencoder is a separate network that needs to be trained independently from the segmentation network. Similarly, Jeya Maria Jose Valanarasu \etal~\cite{valanarasu2022fly} also propose to train an additional autoencoder to adapt the Adaptive Instance Norm (AdaIN) layers, which profoundly relies on the training performance of the additional autoencoder network. 
In contrast, our work utilizes a single network, simplifying the training process and ensuring consistent and reliable results. 

\section{Methodology}

In this section, we first provide an overview of the SDG problem and our proposed DeY-Net. We then introduce the basic principle of the self-supervised denoising scheme Noise2Void~\cite{krull2019noise2void} as the preliminary of DeY-Net and explain denoising training and test-time adaptation (DeTTA) in detail. 

\subsection{Overview}
In the setting of single domain generalization (SDG)\cite{wang2022generalizing}, we are given only one training (source) domain and we denote it as $S_{train}=\{(\textbf{x}_i,y_i)\}_{i=1}^n\thicksim{P_{XY}}$, where $\textbf{x} \in \mathcal{X} \subset \mathbb{R}^d$ denotes the input, $y \in \mathcal{Y} \subset \mathbb{R}$ denotes the label, and $P_{XY}$ denotes the joint distribution of the input sample and output label. $X$ and $Y$ denote the corresponding random variables. 
The goal of SDG is to learn a robust and generalizable predictive function $h$ : $\mathcal{X} \to \mathcal{Y}$ from the single source domain to achieve a minimum prediction error on an unseen test domain $S_{test}$ (i.e., $S_{test}$ cannot be accessed in training and $P_{XY}^{test} \neq P_{XY}^{train}$):
\begin{equation}
    \mathop{\min}\limits_{h}\mathbb{E}_{(x,y)\in S_{test}} [l(h(\textbf{x}), y)], \label{loss}
\end{equation}
where $\mathbb{E}$ is the expectation and $l(\cdot, \cdot)$ is the loss function.

On this basis, test-time adaptation (TTA) utilizes the unlabeled test sample $\textbf{x}_i^{test} \in S_{test}$ presented at test time to adapt the model for generalization purpose.

To address the SDG problem, an overview of our method DeY-Net is illustrated in \cref{fig:Overview}. While we build upon the U-Net architecture, our approach utilizes a Y-shaped design, where the encoder is followed by two decoders: the segmentation decoder for pixel-wise segmentation and the denoising decoder for self-supervised denoising. 

\begin{figure}[th]
\vspace{-4mm}
\includegraphics[width=1\linewidth, trim=0cm 7cm 0cm 0cm, clip]{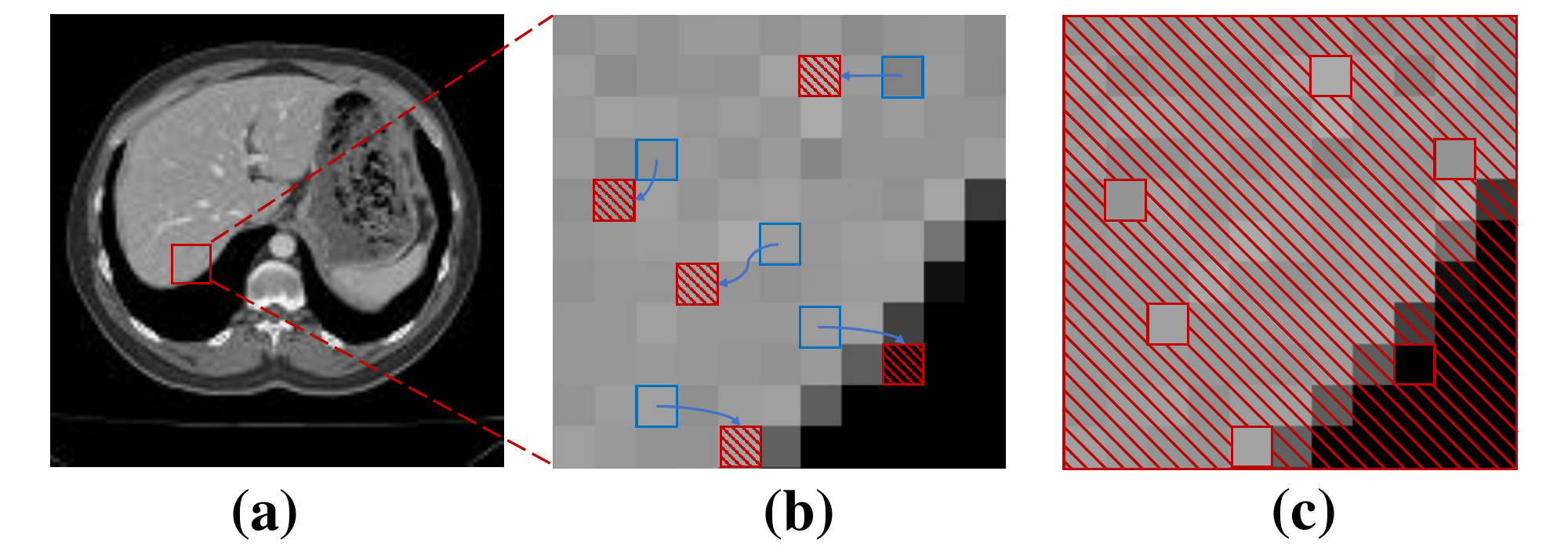}
\vspace{-6mm}
   \caption{
   The training scheme of Noise2Void.
   \textbf{(a)} An original training image. \textbf{(b)} A magnified image patch extracted from (a). Similar operations are performed throughout the entire image. During N2V training, several random pixels (red and striped squares) are replaced by neighboring pixels (blue squares). This modified image is then used as the input image. 
   \textbf{(c)} The target patch corresponding to (b).
   The loss is only calculated for the pixels masked in (b).}
\label{fig:N2V}
\vspace{-5mm}
\end{figure}

\subsection{Preliminary of DeY-Net}

The core idea of DeY-Net revolves around utilizing self-supervised denoising to enhance generalization performance. Before delving into the details of our approach, the principles of the self-supervised denoising will be introduced.
We use the Noise2Void (N2V) scheme described in ~\cite{krull2019noise2void} as our self-supervised denoiser of choice.
Conveniently, N2V uses a default U-Net with a modified input and loss for denoising. We replicate the decoder of the original U-Net to serve as the second decoder for N2V.

In N2V, the noise is assumed pixel-wise independent. Thus the noise information is not carried in the neighboring pixels. Hence, denoising is feasible by predicting the pixels' values replaced by neighboring pixel values. As the training scheme shown in \cref{fig:N2V}, N2V randomly selects $N$ pixels in each training image; then these pixels are replaced with neighboring pixels. The training targets are the corresponding original pixel values.

The N2V network can be trained by minimizing the empirical risk

\begin{equation}
    \mathop{\arg\min}\limits_{f}\sum_{j}\sum_{i}L(f(\tilde{x}_{RF(i)}^j),x_i^j), \label{N2V}
\end{equation}
where $\tilde{x}_{RF(i)}^j$ is a patch around pixel $i$, extracted from training input image $x^j$. In this patch, the value at the position $i$ is replaced with the value of a neighboring pixel. $x_i^j$ is the corresponding target pixel value. The summation of the losses of all $N$ pixels $i$ in the entire training image yields the total loss for each image.

For $L$, we consider the standard MSE loss:
\begin{equation}
    L(\tilde{s}_i^j,s_i^j) = (\tilde{s}_i^j-s_i^j)^2.
\end{equation}

\subsection{Training of DeY-Net}

\textbf{Pretraining.} To utilize the efficacy of denoising, we perform a separate pretraining step for the segmentation decoder using a well-trained denoising network~\cite{brempong2022denoising}. Specifically, we trained a U-Net with the Noise2Void training scheme, utilizing all the source domain data without any segmentation labels. Subsequently, we copy the parameters of the trained decoder to the segmentation decoder of DeY-Net, while the encoder and the denoising decoder of DeY-Net are randomly initialized.
If we simultaneously pretrain other components, it may result in overfitting. In the ablation experiments, we will demonstrate the effectiveness of the pretraining step and compare the results obtained by pretraining different components separately.

\textbf{Joint-training.}  \label{joint}
As shown in \cref{fig:Overview}, our Y-shaped architecture allows the simultaneous execution of two tasks. We train both the segmentation and the denoising tasks by summing their respective losses.
The encoder, the pretrained segmentation decoder, and the denoising decoder are denoted as $f$, $s_0$, and $d$, respectively.

We randomly mask pixels of the input image $x^j$ from the training set and replace them with the neighboring pixel values, getting the actual input $\tilde{x}^j$. The original input values at the masked positions are the training targets for the denoising task. 

The denoising loss is evaluated for labeled images and unlabeled ones. We use the standard Noise2Void loss, which is expressed as:
\begin{equation}
    l_{de}=\sum_j\sum_iL(d\circ{f(\tilde{x}_i^j)},x_i^j). \label{l_de}
\end{equation}
To address the common imbalance in the number of foreground and background pixels processed in medical images, we choose the standard Dice loss~\cite{blanchard2021domain} as the supervised segmentation loss, evaluated for labeled images only, with labels as $\textbf{y}$. The segmentation loss can be expressed as:
\begin{equation}
    l_{seg}= \sum_jDice(s_0\circ{f(\tilde{x}^j)},y^j). \label{l_seg}
\end{equation}
In contrast to the straightforward joint-training process in TTT~\cite{sun2020test}, our method incorporates an enhanced joint-training process by introducing a time-dependent weight to better combine supervised and unsupervised loss~\cite{laine2016temporal}. The joint-training produces a trained encoder $f_0$ and two trained decoders $s_1$ and $d_0$:
\begin{equation}
    f_0,d_0,s_1=\mathop{\arg\min}\limits_{f,d,s}(l_{seg}+w(t)*l_{de}).
\end{equation}
In our implementation, the weighted function $w(t)$ of unsupervised loss slopes upwards from 0 along the Gaussian curve for the first 200 training periods. It means that the denoising decoder, which assists the encoder in feature extraction, slowly starts to work during the training process. Initially, the model training is primarily driven by the segmentation loss, ensuring the model does not converge to a degenerate solution where meaningful segmentation is not achieved~\cite{laine2016temporal}.
This weight adjustment allows the main segmentation task and the auxiliary denoising task to strike a balance, ensuring practical completion of the segmentation task while preserving the generalization ability offered by the denoising task.

\subsection{DeTTA}

\textbf{Target domain adaptation.} At test time, considering that the single test volume can give us a hint about its distribution, we aim to optimize the model parameters with each unlabeled test volume. Once each test volume $x$ arrives, we optimize the denoising loss (\cref{l_de}) on the denoising branch $d_0\circ{f_0}$, while the segmentation decoder $s_1$ is frozen:
\begin{equation}
    f_x,d_x=\mathop{\arg\min}\limits_{f,g}L(d_0\circ{f_0(\tilde{x})},x). \label{adapt}
\end{equation}

We only perform a one-step gradient descent on each test volume, a medical volume data from a specific patient. Using the whole volume of data simultaneously is consistent with the clinical scenario where test data usually arrives per patient.
To preserve the original discriminability of the model, we only adapt the parameters in the batch normalization layers for the test-time adaptation.
It is motivated by the fact that modifying all the parameters of the model is unstable and inefficient when only a single test sample is available at test time~\cite{wang2020tent}.

After test-time adaptation to each test volume, we get an adapted model for segmentation $s_1\circ{f_x}$, and we make predictions on the test volume $x$ as $s_1\circ{f_x(\tilde{x})}$. 
Note that we use $\Tilde{x}$ instead of $x$ as the prediction input.

The above adaptation in our method is not performed online, as we do not assume that each medical volume data comes from the same distribution.
After making predictions on each test volume $x$, we always discard $f_x$ and $d_x$, and reset the weights to $d_0$ and $f_0$ for the next test volume.

\textbf{Noise-corrupted input adaptation.} Indeed, it is apparent that the modified input $\tilde{x}$ leads to the loss of information of the original test image. As mentioned in \cref{joint}, the inputs during training are modified as $\tilde{x}$ to facilitate the joint-training of the two tasks. If we directly make predictions on original $x$ as $s_1\circ{f_x(x)}$ to preserve all the original information, the reliability of the predictions may be compromised.
To further preserve the information, DeY-Net further adapts to the noise-corrupted input.

\begin{figure}[th]
\vspace{-3.5mm}
\includegraphics[width=1\linewidth, trim=2cm 4.2cm 3cm 1.1cm, clip]{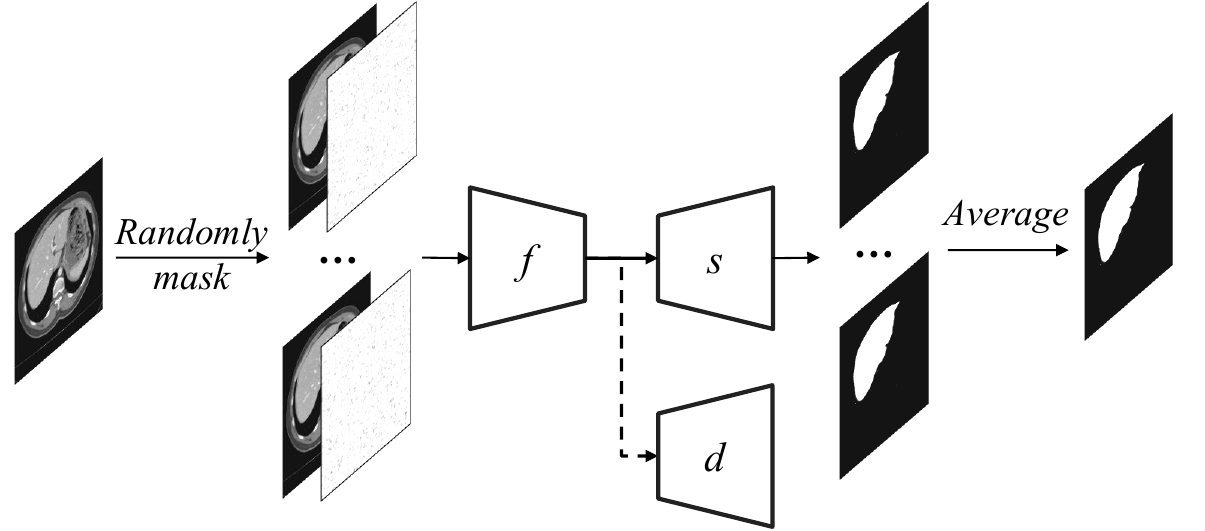}
\vspace{-6mm}
   \caption{The pipeline of noise-corrupted input adaptation.}
\label{fig:Train}
\vspace{-5mm}
\end{figure}

Intuitively, for each original test $x$, we can get several inputs $\{\tilde{x},\tilde{x}^{'},\tilde{x}^{''},\cdots\}$ after being randomly masked separately. 
Since the positions of the masked pixels are random, such a set of inputs minimizes the information loss. Then, we make predictions on all these masked inputs and merge the predictions. It is a novel test-time augmentation strategy designed for our method. The pipeline is shown in \cref{fig:Train}.

\section{Experiment}
We assess the effectiveness of the proposed DeY-Net and DeTTA on a typical medical image segmentation: liver segmentation on CT. Experimental details, comparison results with other methods, and ablation studies are detailed in the following subsections.

\subsection{Experimental settings}

\textbf{Datasets.} 
We use CHAOS-train~\cite{kavur2021chaos} with 20 CT volumes as the labeled dataset and CHAOS-test~\cite{kavur2021chaos} with 20 CT volumes as the unlabeled dataset. Both the labeled and unlabeled data are collected from healthy patients only. We further randomly split the CHAOS-train dataset into train and test sets with 16 and 4 volumes, respectively, for source domain training and in-domain testing. To assess the out-of-domain generalization, we evaluate three additional out-of-domain datasets, including LITS2017~\cite{bilic2023liver} and two datasets from local clinical centers, named Normal and Ill. LITS2017 is a challenging dataset that contains 130 CT volumes from healthy patients and patients with liver tumors. Normal contains 30 CT volumes from healthy patients, while Ill contains 15 from patients with cirrhosis, both annotated by experts.

\begin{figure}[th]
\begin{center}
\vspace{-2mm}
\includegraphics[width=0.9\linewidth, trim=1.5cm 2.8cm 1cm 3.5cm, clip]{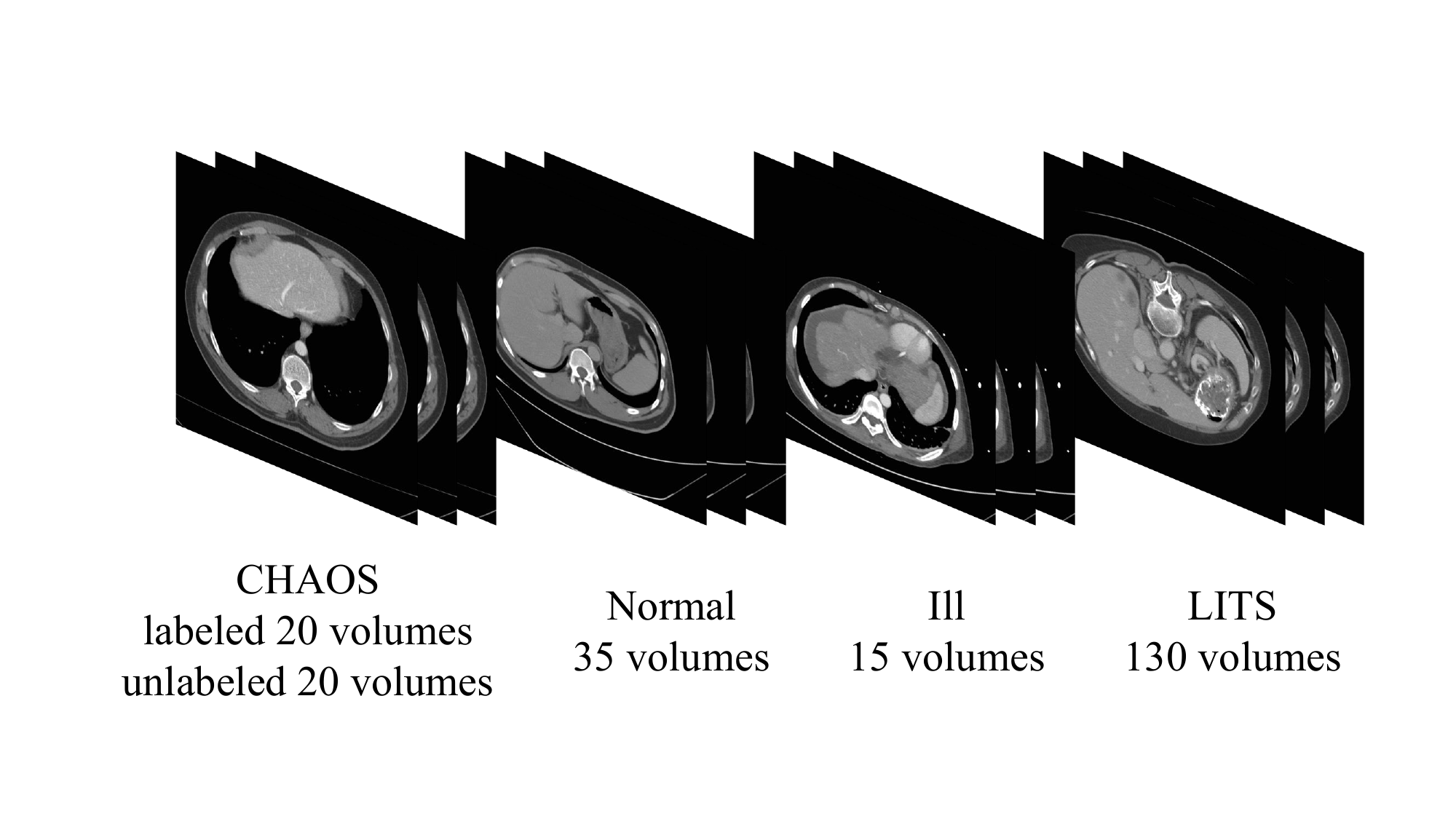}
\vspace{-2mm}
   \caption{The representative cases and the volume number of 4 datasets.}
\label{fig:dataset}
\vspace{-7mm}
\end{center}
\end{figure}

These datasets constitute at least 4 data domains, of which LITS2017 collected by multiple clinical centers, simply handled as a data domain. \cref{fig:dataset} shows the representation cases and volume number of each dataset, showing the domain differences (such as the liver’s position, resolution, and direction). In addition, liver morphology varies between patients with cirrhosis, tumors, and healthy patients. Since the thickness of data slices varies greatly between clinical sites, we validated our method on these data using a 2D network as the backbone network.

\begin{figure*}
\footnotesize
    \begin{minipage}[t]{0.65\linewidth}
        \centering
        \begin{tabular}{c|c|cccc}
            \hline\hline
            \multirow{2}*{\textbf{Dataset}} & in-domain & \multicolumn{4}{c}{out-of-domain} \\
            \cline{2-6}
            & CHAOS test & Normal & Ill & LITS & average \\
            \hline\hline
            & \multicolumn{5}{c}{\textbf{Dice Coefficient↑ (Dice, mean±std)}} \\
            \hline\hline
            U-Net~\cite{ronneberger2015u} & 96.32±0.20 & 92.71±0.90	& 88.75±0.51	& 85.33±0.29 &	90.78 \\
            \hline
            BigAug~\cite{8995481} & 96.34±0.05 & 93.08±0.18 & 89.21±0.80 & 85.28±0.31 & 90.98 \\
            DualNorm~\cite{Zhou_2022_CVPR} & 96.20±0.16 & 94.09±0.03 & 90.22±0.28 & 83.54±0.23 & 91.01 \\
            TTT~\cite{sun2020test} & 95.62±0.43 & 94.57±0.31 & 90.54±0.40 & 84.45±0.27 & 91.61 \\
            Tent~\cite{wang2020tent} & 96.55±0.04 & 95.24±0.04	& 90.79±0.46 & 85.55±0.31 & 92.03 \\
            RN+CR~\cite{hu2021fully} & 96.59±0.08 & 95.26±0.04 & 90.76±0.47 & 85.58±0.33 & 92.05 \\
            TTST~\cite{karani2021test} & 96.71±0.17 & 95.32±0.33 & 91.48±0.26 & 85.67±0.66 & 92.30 \\
            OtF~\cite{valanarasu2022fly} & 96.22±0.01 & 94.45±0.03 & 88.14±0.08 & 81.17±0.01 & 90.00 \\
            \hline
            ReY-Net (\textit{w/o} ReTTA) & 96.57±0.06 & 95.72±0.04	& 90.44±0.51 & 82.35±2.76 & 91.27 \\
            ReY-Net (\textit{w/} ReTTA) & 96.23±0.95 & 95.51±0.16	& 90.93±0.26 & 86.34±0.36 & 92.26 \\
            \hline
            DeY-Net (\textit{w/o} DeTTA) & 96.63±0.03 & \textbf{95.74±0.04}	& 91.50±0.18 & 85.19±0.04 & 92.27 \\
            DeY-Net (\textit{w/} DeTTA) & \textbf{96.71±0.12} & 95.66±0.08 & \textbf{91.60±0.14} & \textbf{87.14±0.09} & \textbf{92.77} \\
            \hline\hline
        \end{tabular}
        \captionof{table}{Quantitative comparison of domain generalization results.}
        \label{table:result}
    \end{minipage}%
    \hfill
    \begin{minipage}[t]{0.33\linewidth}
        \begin{minipage}[t]{1\linewidth}
            \centering
            \includegraphics[width=1\linewidth, trim=0cm 0.3cm 0cm 0.3cm, clip]{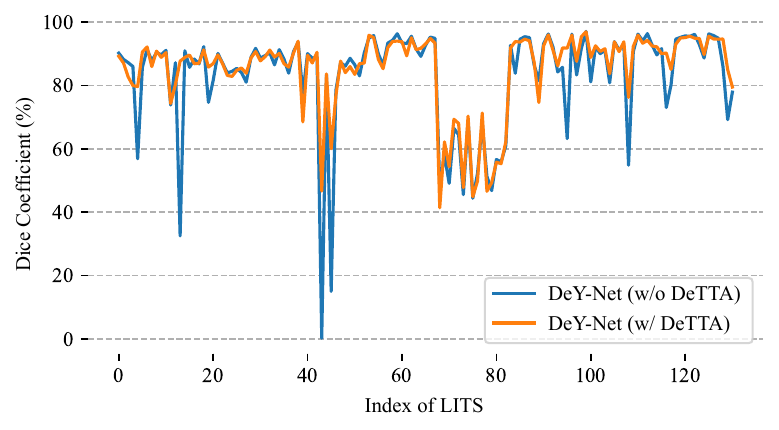}
            \vspace{-6mm}
            \caption{DeTTA improvement of LITS.}
        \label{fig:analysis}
        \end{minipage}
        \vfill
        \begin{minipage}[t]{1\linewidth}
            \centering
            \begin{tabular}{c|cc}
            \hline\hline
            \textbf{Dataset}& LITS & Tumor \\
            \hline\hline
            DeY-Net (\textit{w/o} DeTTA) &  85.19 & 83.92\\
            DeY-Net (\textit{w/} DeTTA) & 87.14 & 88.51 \\
            \hline
            $\triangle$ & \textbf{1.95} & \textbf{4.59} \\
            \hline\hline
            \end{tabular}
            \vspace{-2mm}
            \captionof{table}{Quantitative results of DeTTA improvement. Tumor datasets consists of images containing tumors in the LITS dataset.}
        \label{table:DeTTA}
        \end{minipage}
    \end{minipage}
    \vspace{-5mm}
\end{figure*}
\textbf{Preprocessing.} 
LITS2017 is a collection of liver CTs with liver and tumor segmentation labels. We only use the liver class and merge the tumor class into the liver class.
The preprocessing in ~\cite{li2021semantic} is used for all the raw CT data. We first clip the attenuation coefficient in a range between -200 and 400, highlighting the liver portion, and subsequently normalize it by subtracting the minimum and dividing by the signal range in a slice-wise manner. 
To process the input, we set the masked ratio at 0.1, meaning that 10\% pixels of each input are selected and replaced with the neighboring pixel values following N2V. 
We do not do any cropping, resampling, or alignment but only the slice-wise preprocessing described above.

\textbf{Metrics.} 
For evaluation, we use a commonly used metric, the Dice coefficient [\%]~\cite{dice1945measures}, to quantitatively evaluate the segmentation results.
To ensure the stability and reliability of our experiments, we conduct each experiment three times utilizing different random seeds.
The results are presented using the average value with the standard deviation.


\subsection{Implementation details}
In our experiment, the test-time adaptation updates just a step with a learning rate of 1e-6. For the test-time augmentation, we augment twice and average the two outputs. 
We use the U-Net as the backbone for all experiments. No data augmentation technology is utilized in our methods and other comparison methods.
In our method, the U-Net is also used for the pretrained denoising model, which is trained with all the CHAOS data of 200 epochs. Both the pretrained denoising U-Net and our proposed model DeY-Net are trained using Adam optimizers~\cite{kingma2014adam} with the momentum of 0.9 and 0.99, and the learning rate is initialized to 1e-4. We train the DeY-Net for 200 epochs on the CHAOS dataset only until convergence. The batch size of the baseline U-Net and the DeY-Net are consistently set to 8, which means four labeled data and four unlabeled data in each batch of DeY-Net.
During training, the weighted function $w(t)$ of unsupervised loss slopes upwards from 0 along the Gaussian curve for the first 200 training periods. The maximum value of $w(t)$ is $\alpha({n_l}/({n_l}+n_{un}))$, where $n_l$ and $n_{un}$ are numbers of labeled and unlabeled data, respectively. After conducting exploratory experiments, we set the $\alpha$ to 30 in our experiments.
The framework is implemented via Pytorch using an NVIDIA P100 GPU.

\subsection{Results}

\textbf{Comparison methods.}
We compare our methods with current state-of-the-art methods to solve the SDG problem, including: 
\textbf{BigAug (2020)}~\cite{8995481}, a DG method in medical image segmentation with extensive data transformations to promote general representation learning.
\textbf{DualNorm (2022)}~\cite{Zhou_2022_CVPR}, an SDG method in medical image segmentation via style augmentation and dual normalization.
\textbf{TTT (2020)}~\cite{sun2020test}, a test-time training method in image classifier with an auxiliary self-supervised task of rotation prediction. 
\textbf{Tent (2021)}~\cite{wang2020tent}, a fully test-time adaptation method in the image classifier field by minimizing the entropy of predictions. 
\textbf{TTST (2021)}~\cite{karani2021test}, a test-time adaptation in medical image segmentation with an additional denoising autoencoder. 
\textbf{RN+CR (2021)}~\cite{hu2021fully}, a fully test-time adaptation method in medical image segmentation by minimizing new losses, Regional Nuclear-norm (RN) and Contour Regularization (CR). 
\textbf{OtF (2022)}~\cite{valanarasu2022fly}, an on-the-fly test-time adaptation method with an additional Domain Prior Generator. 

The \textbf{Baseline} setting in domain generalization denotes learning a model (U-Net~\cite{ronneberger2015u}) on the source domain without using any generalization technique and directly making predictions on the target domains.

\textbf{Quantitative comparison results.}
\cref{table:result} shows the results on the liver segmentation. For the comparison methods using TTA, we also perform a one-step gradient descent for each testing volume. 
First, we compare our method DeY-Net (\textit{w/} DeTTA) against the baseline U-Net. While the U-Net achieves satisfactory in-domain performance but struggles with out-of-domain data, our proposed method further enhances in-domain performance and significantly improves performance on unseen domains, with an improvement of 1.99\% on average.

Most of the methods can improve the generalization performance over baseline.
Among all these methods, our DeY-Net (\textit{w/} DeTTA) performs better than other methods. 
One noticeable observation is the significant improvement achieved by our method compared to OtF. Due to removing the back-propagation, OtF heavily relies on the generalization performance of the additional Domain Prior Generator, highlighting our joint-training approach's advantage.
Compared with BigAug and DualNorm, the DG methods that rely on data augmentation, and TTT and Tent, the TTA methods that are not designed for the image segmentation task, our method has a more significant improvement, especially on the most challenging LITS dataset. Even though RN+CR, a method similar to Tent, is designed for image segmentation, only a slight improvement over Tent is achieved in the liver segmentation experiments.

Notably, our method demonstrates the slightest improvement over TTST. 
This can be attributed to using an additional denoising autoencoder trained outside the segmentation network in TTST to refine the segmentation results. The sequential structure connection between the two networks consumes computational resources and adds complexity to the overall process.
Instead of the sequential structure in TTST, we employ an alternative and more convenient Y-shaped architecture for leveraging self-supervised denoising, which has resulted in superior performance.

Within these out-of-domain datasets, the results of LITS are more representative of generalization ability, which contains the largest and most diverse data. These comparison methods fail in LITS dataset on average. 
Meanwhile, our method still improves performance on the LITS dataset, demonstrating its effectiveness even in challenging unseen domains.

\begin{figure}[t]
\centering
\vspace{-4mm}
    \includegraphics[width=1\linewidth, trim=1.4cm 1.5cm 2cm 0.5cm, clip]{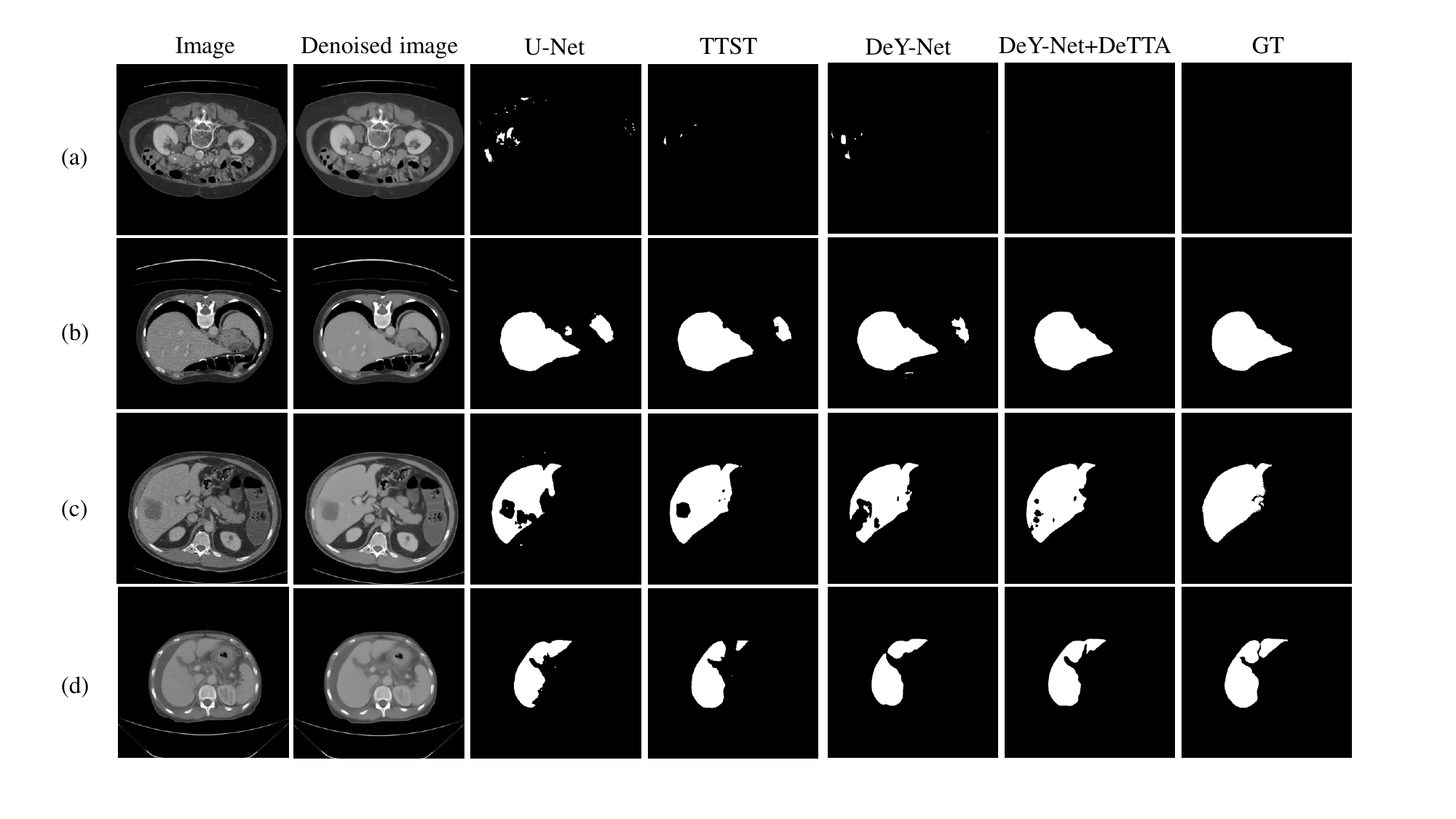}
    \vspace{-5mm}
   \caption{
       Visualization comparison of segmentation results with baseline methods, TTST and the proposed method DeY-Net.
       The samples represent the following:
        \textbf{(a)} A liver-free image;
        \textbf{(b)} A healthy liver image;
        \textbf{(c)} An image with liver tumors;
        \textbf{(d)} An image with liver cirrhosis.
    }
\label{fig:result}
\vspace{-6mm}
\end{figure}

\textbf{Visualization comparison results.}
\cref{fig:result} further shows the segmentation results with four samples from unseen domains for the liver segmentation task. It is observed that our method demonstrates superior segmentation accuracy in unseen domains, whereas other methods may fail at times. 
Our method accurately segments the regions as non-liver, which are erroneously segmented as liver by other methods. Furthermore, for the regions containing the liver, our approach captures finer details and accurately delineates the liver's boundaries.
In the case of CT images from patients with liver cirrhosis and liver tumors, all methods struggle to completely overcome domain shifts caused by changes in liver texture and structure. Nonetheless, our method consistently achieves more reliable segmentation results, benefiting from its effective image information utilization. This improvement illustrates the role of the denoising branch in capturing intricate details of the images.


\textbf{Analysis of DeTTA improvement.} \label{TTA}
Overall, DeTTA increases the average Dice by 0.50\% over DeY-Net(\textit{w/o} DeTTA), with a notable increase of 1.95\% in the LITS dataset. 
This observation reflects that DeTTA gains additional capacity and enables the model to utilize the test data information to improve model generalizability adaptively.

However, TTA methods are not always work due to varying degrees of domain shifts~\cite{yang2022dltta}. Analyzing scenarios in which DeTTA encounters limitations is crucial.
Our detailed analysis, conducted primarily on the challenging LITS dataset (\cref{fig:analysis} and \cref{table:DeTTA}), reveals that for certain test data with large domain shift and poor segmentation results, DeTTA can extract information from the test data and adapt the model according to the test domain to significantly improve the segmentation results. This is quantitatively demonstrated by the improvement of Tumor dataset(4.59\%), a subset of the LITS dataset(1.95\%) with more pronounced domain shifts.
Conversely, when the initial segmentation results are already satisfactory, indicating a slight shift between the test data and the source domain distribution, employing DeTTA may yield slight improvement or even lead to a decline in performance.
We attribute the observed decline to the potential over-optimization of the trained model to the tested sampling, leading to performance degradation even with a single step of gradient update. 
We anticipate that this can be decently resolved by an adaptive DeTTA strategy, which is beyond the scope of this work and left for future exploration.

\subsection{Ablation study}

We conduct ablation studies about several key points in our model: (1) the contribution of DeTTA in our method; (2) the effect of the weighted function $w(t)$; (3) the effect of the optimized layers; (4) the effect of denoising pretraining; (5) the effect of data augmentation times.


\textbf{Effect of denoising task.} \cref{table:result} shows the effect of different self-supervised tasks on the final results. TTT employs image rotation prediction as the self-supervised task. ReY-Net extends from our DeY-Net (with consistent network architecture, training techniques, etc.). and the self-supervised task strategically opts for the reconstruction most pertinent to denoising. ReTTA, accordingly, employs the reconstruction loss as its optimization objective.
The outcomes demonstrate that self-supervised denoising is more conducive than alternative approaches for the main task of medical image segmentation.

\begin{figure}[t]
\includegraphics[width=1\linewidth, trim=0cm 0.3cm 0cm 0.1cm, clip]{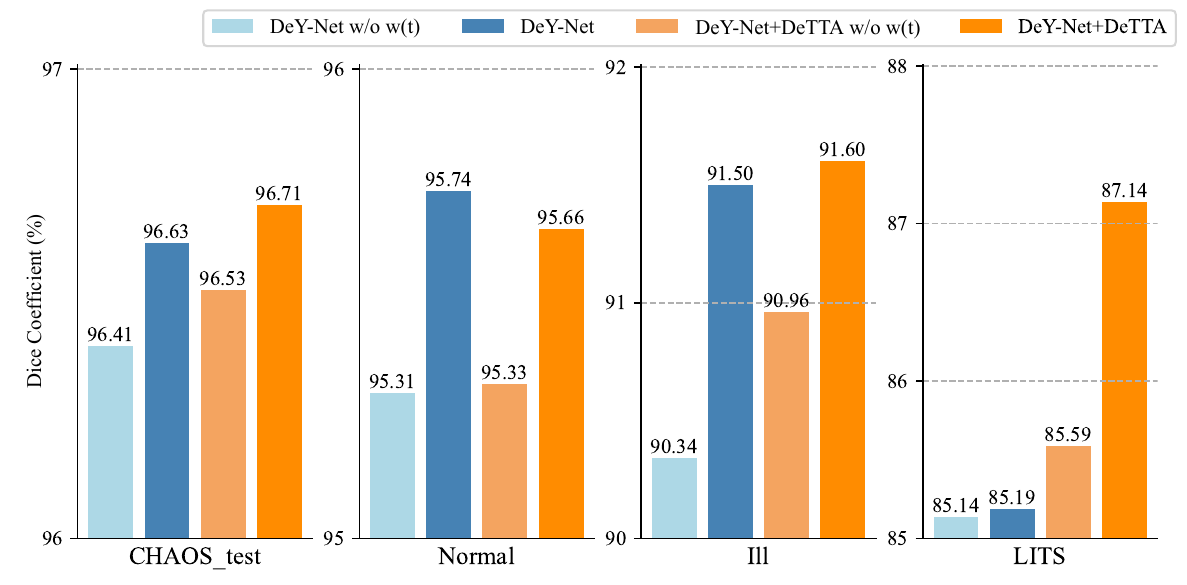}
\vspace{-7mm}
   \caption{
   Ablation analysis of time-dependent weight.
   We analyze the effectiveness of the time-dependent weight on models with and without DeTTA.
   }
\label{fig:w_t}
\vspace{-3mm}
\end{figure}

\textbf{Effect of $w(t)$.} As shown in \cref{fig:w_t}, $w(t)$ is effective for performance improvement of DeY-Net, which means that gradually increasing the supervision of unsupervised loss during the training process is beneficial. Notably, no matter how well the DeY-Net is trained, after the DeTTA, the overall performance will still be improved.

\textbf{Effect of the optimized layers.} 
We explore which layers need to be adapted when a model is transferred to the unseen domain. In the DeTTA, only the encoder can be optimized for segmentation, and the segmentation decoder is frozen. As shown in \cref{table:layers}, adapting all encoder parameters is worse than adapting BN layers. It is intuitive that all parameters are adapted easily to damage the original performance of the network due to the network being over-parameterized.

\begin{table}
\begin{center}
\begin{tabular}{c|cccc}
\hline\hline
\textbf{Optimized layers} & Normal & Ill & LITS  \\
\hline\hline
BN & \textbf{0.9566} & \textbf{0.9160} & \textbf{0.8714} \\
All & 0.9566 & 0.9159 & 0.8696 \\
\hline\hline
\end{tabular}
\end{center}
\vspace{-6mm}
\caption{Ablation analysis of the optimized layers.}
\label{table:layers}
\vspace{-5mm}
\end{table}

\textbf{Effect of denoising pretraining. } 
As shown in \cref{table:pretrain}, we conduct comprehensive experiments regarding the initialization methods for DeY-Net. The results represent the average Dice score across the four datasets.
The (3) initialization method performs best when only the segmentation decoder is pretrained using a denoising U-Net. 
This particular pretraining approach outperforms other pretraining methods since joint-training is one key aspect of our approach. Exploiting the overlap between segmentation and denoising tasks, we pretrain the segmentation decoder to ease the segmentation task's complexity. Meanwhile, we refrain from pretraining the denoising branch to prevent premature convergence, enabling more effective joint-training of the two tasks.
Compared to the random initialization (1), pretraining the segmentation decoder allows for better utilization of the denoising parameters trained on labeled and unlabeled data.




\begin{table}
\begin{center}
\footnotesize
\begin{tabular}{c|c|c|c|c|c}
\hline
 idx & $f$ & $s$ & $d$ & \textit{w/o} DeTTA & \textit{w/} DeTTA \\
\hline
(1) & \ding{55} & \ding{55} & \ding{55} & 0.9169 & 0.9246 \\
(2) & \ding{51} & \ding{55} & \ding{55} & 0.9154 & 0.9211 \\
\textbf{(3)} & \ding{55} & \ding{51} & \ding{55} & \textbf{0.9227} & \textbf{0.9277} \\
(4) & \ding{55} & \ding{55} & \ding{51} & 0.9206 & 0.9215 \\
(5) & \ding{51} & \ding{51} & \ding{55} & 0.9177 & 0.9208 \\
(6) & \ding{51} & \ding{55} & \ding{51} & 0.9154 & 0.9170 \\
(7) & \ding{55} & \ding{51} & \ding{51} & 0.9183 & 0.9233  \\
(8) & \ding{51} & \ding{51} & \ding{51} & 0.9084 & 0.9185 \\

\hline
\end{tabular}
\end{center}
\vspace{-6mm}
\caption{Ablation analysis of the pretraining. The encoder, the segmentation decoder, and the denoising decoder are denoted as $f$, $s$, and $d$, respectively. \ding{51} represents pretraining, while \ding{55} represents random initialization. Our pretraining method is (3), while the baseline is (1).} 
\label{table:pretrain}
\vspace{-1mm}
\end{table}

\begin{figure}[t]
\begin{center}
\vspace{-1mm}
\includegraphics[width=0.8\linewidth, trim=0cm 0cm 0cm 0.2cm, clip]{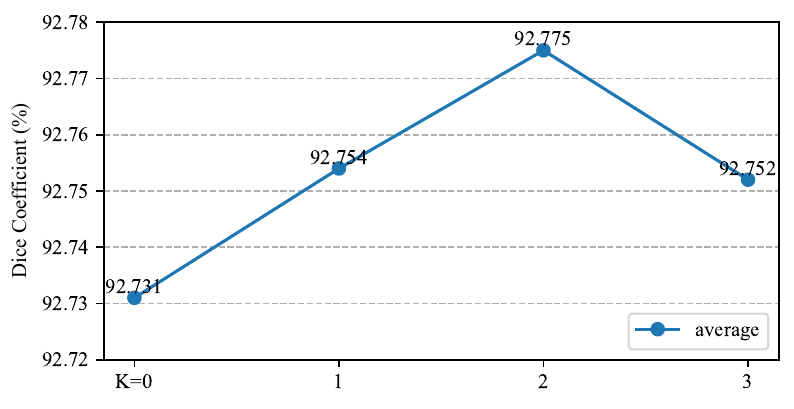}
\end{center}
\vspace{-8mm}
   \caption{Ablation analysis of data augmentation times $K$.}
\label{fig:average}
\vspace{-5mm}
\end{figure}

\textbf{Effect of data augmentation times.}
The choice of data augmentation times $K$ is important in our method, affecting both the final performance and the testing efficiency. To investigate the suitable choice of $K$, we repeat the experiment of the DeY-Net by varying $N\in(0,1,2,3)$. $K=0$ represents the original input without being randomly masked. $K=1$ represents the scenario without adaptation to the noise-corrupted input (data augmentation). As shown in \cref{fig:average}, the models with data augmentation times (K=2) perform better than those with smaller or larger times on the segmentation task. We finally adopt K=2 in our method.

\section{Conclusion}
This paper proposes Denoising Y-Net (DeY-Net) to address the challenging SDG problem in medical image segmentation. The idea is to incorporate an auxiliary denoising decoder into a basic U-Net architecture, that naturally allows for semi-supervised training and shows strong generalization capabilities. Further, we propose Denoising Test Time Adaptation (DeTTA) to adapt the model to the target domain and adapt to the noise-corrupted input, which can further promote the model generalization at any unseen data distributions. We validate our method in the liver segmentation task. Quantitatively, we significantly outperform our baseline and other methods in- as well as out-of-domain. 
DeTTA may over-optimize the test data in some scenarios, leading to performance degradation even with a single step of gradient update. Also, the network architecture of our method is simple and can cope well with the domain generalization problem in the same modality case. Therefore, the future direction of research lies in developing an adaptive DeTTA strategy and cross-modality generalization.

\section*{Acknowledgement}

This research was supported by the Horizontal Project of Zhejiang University (K20220130).


\newpage
{\small
\bibliographystyle{ieee_fullname}
\bibliography{egbib}
}
\end{document}